# Hand Gesture Recognition Library

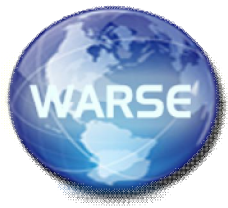

Jonathan Fidelis Paul[1], Dibyabiva Seth[2], Cijo Paul[3], Jayati Ghosh Dastidar[4]
[1] St. Xavier's College, Kolkata, India, jfdpaul@gmail.com
[2] St. Xavier's College, Kolkata, India, meetdseth@gmail.com
[3] St. Xavier's College, Kolkata, India, pl.cijo@gmail.com
[4] St. Xavier's College, Kolkata, India, j.ghoshdastidar@sxccal.edu

**ABSTRACT**

In this paper we have presented a hand gesture recognition library. Various functions include detecting cluster count, cluster orientation, finger pointing direction, etc. To use these functions first the input image needs to be processed into a logical array for which a function has been developed. The library has been developed keeping flexibility in mind and thus provides application developers a wide range of options to develop custom gestures.

## 1. INTRODUCTION

Gestures are closely tied to humans both in interpretation and expression. Gestures are mostly space-bound and human gestures include a high percentage of hand usage. Since hand gestures are so closely tied to us, thus the motivation to use gestures to control technology, which has become an integral part of our living, is justified.

For a hand gesture recognition library it is a prerequisite to be able to detect the states of the hand, its position and track it. It should also comprise functions to handle various Operating system processes.

In this paper we have presented a library to recognize hand gestures. It includes a variety of functions from counting the number of clusters, to finding its orientation and tracking the position of each disconnected cluster. These functions can be employed by an application developer in conjunction, to alleviate the use of keyboards and mouse to generate gestures that can almost substitute traditional input device use.

In the following sections we have explained the working of the various algorithms we have used to create the proposed library.

## 2. LITERATURE REVIEW

Hand Gesture Recognition is a field which has been explored by academicians and scientists the world over. Literature Review suggests that authors have used hidden Markov models, 2D Motion trajectories, and other allied areas to tackle the problem.

Authors in [1] have presented algorithms for hand-tracking and hand-posture recognition in terms of multi-scale color images.

The authors in paper [2] have used real-time tracking and hidden Markov models to trace the moving hand and extract the hand region. They then used the Fourier descriptor to characterize spatial features.

A scheme for extracting and classifying two-dimensional motion in an image sequence based on motion trajectories was presented by the authors in [3]. They used multi-scale segmentation, affine transformations and time-delayed neural network to show that motion patterns of hand gestures can be extracted and recognized accurately using motion trajectories.

The authors in [4] have successfully patented a technology which makes use of color segmentation, region labeling and principal component analysis.

A glove has been used by the authors in [5] to create a hand-machine interface. The glove is assumed to have been fitted with sophisticated sensors which are capable of sensing flex movements, measure finger bending, etc. Hand position and orientation are measured either by ultrasonics, providing five degrees of freedom, or magnetic flux sensors, which provide six degrees of freedom. The wearer of the glove would thus have an interface to a visual programming language.

In paper [6] the authors have proposed a two level approach to solve the problem of real-time vision-based hand gesture classification. The lower level of the approach implements the posture recognition with Haar-like features and the AdaBoost learning algorithm. The higher level implements the linguistic hand gesture recognition using a context-free grammar-based syntactic analysis. Given an input gesture, based on the extracted postures, the composite gestures can be parsed and recognized with a set of primitives and production rules.

Paul S Heckbert has presented a seed-fill algorithm in [7].

Authors in [8] have presented a modified new seed-fill algorithm.

The authors in [9, 10, 11] have proposed different methodologies to detect and find out whether pixels are aligned in a straight line or not.

Our method makes use of extremely low cost gadgetry which includes LEDs and a glove to implement a library for recognition of hand gestures. These gestures can be used to emulate various mouse operations. The methods used are simplistic, efficient, and easy to implement because it does not employ too many complex operations.

## 3. HAND GESTURE RECOGNITION LIBRARY

### 3.1 Processing Input

It is very essential for the computer to make sense of what it sees through the vision device (like cameras). But the computer cannot do this by itself. Thus in our library we have included a function to process an image as per the developer's choice.

Since colors have various intensities, we have normalized them down to zeroes and ones.

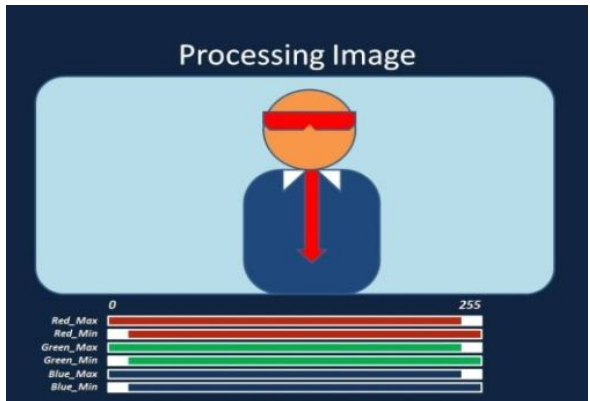

**Figure 1. Input RGB image**

Those colors within the defined range are converted to white (i.e. 1) and the colors not in range are converted to black (i.e. 0). For instance, the spectacles and tie in **Figure 1** have been detected using a color range as shown in **Figure 2.**

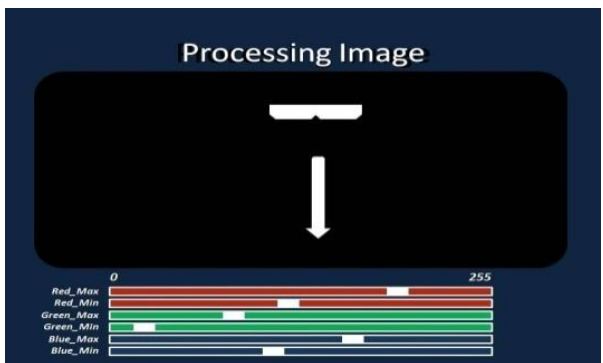

**Figure 2. Output logical array**

To test our library, we have assigned the color red to the fingertips using red LEDs to remove the problems of background noise and poor lighting conditions (**Figure 3).**

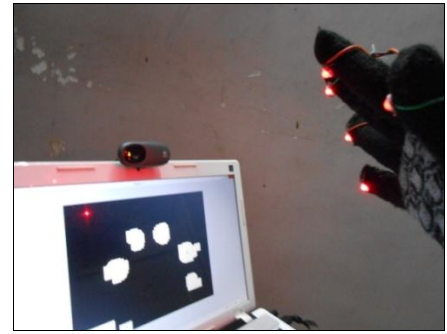

**Figure 3. Demo Application**

### 3.2 Cluster Recognition

In the **hand gesture recognition library**, the concept of cluster extraction has been used.

A **cluster** is a region of connected pixels having the same color values. For example, a set of pixels having intensity of 1 and which are connected directly or indirectly to each other are collectively called a **cluster** (assuming the use of a 2D array).

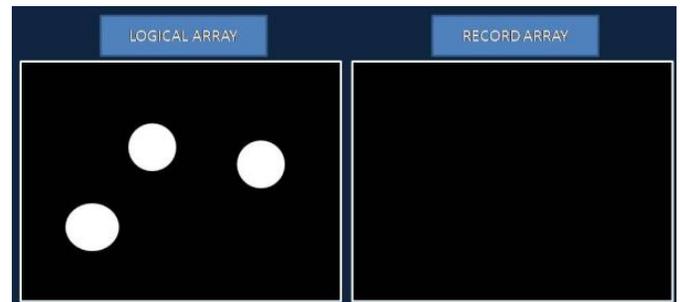

**Figure 4. Original image**

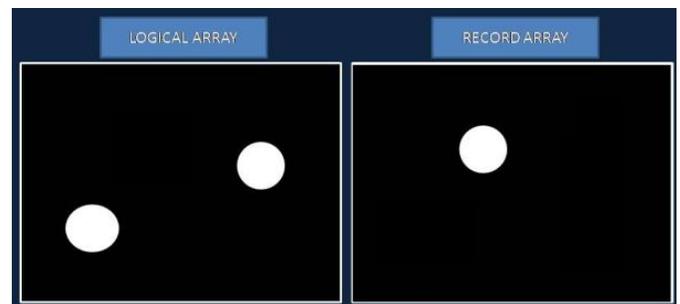

**Figure 5. First cluster extracted**

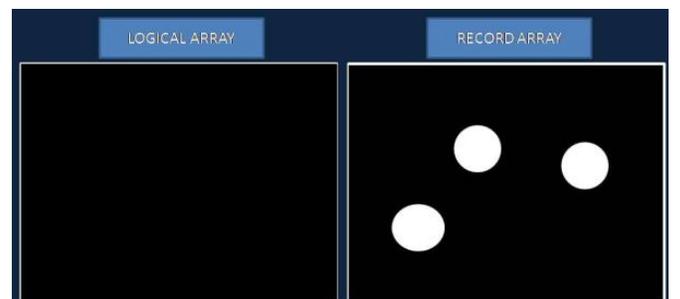

**Figure 6. All clusters extracted**

To extract these clusters we have employed the **seed-fill algorithm** [7, 8] as follows:
The algorithm goes as follows:-
Step 1. Search for the first white pixel in the array
Step 2. If such a pixel is found then push the pixel into the stack and also make the corresponding position of the record_array white, Else goto step 8
Step 3. Increment the counter by 1
Step 4. Pop a pixel from the stack and make the present pixel black.
Step 5. If adjacent pixels are white and not in the record_array (using the **seed-fill algorithm**), then push them into the stack and also make the corresponding entry in the record_array white.
Step 6. If stack is not empty, goto step 4
Step 7. If not end of array, goto step 1
Step 8. End

This algorithm effectively recognizes the clusters and keeps a count of the number of clusters. Since the clusters are found, their positions and numbers too are determined. The number of pixels in each cluster is also determined using this algorithm. (**Figure 4, Figure 5, Figure 6**)

### 3.3 Alignment Detection
Different human gestures have different hand orientations. Thus it is necessary to distinguish between the different hand orientations/alignments [9, 10, 11].
To do this, the simple concept of a straight line alignment was followed. Let us assume that the fingers are represented by single points. Then, if a set of points have the same y-coordinates or x-coordinates then they will be said to be horizontally aligned or vertically aligned respectively. Any pixel with either a different y-coordinate or a different x-coordinate will not be in the same line with the other pixels and hence will not be horizontally or vertically aligned with the others respectively. (**Figure 7, Figure 8**)

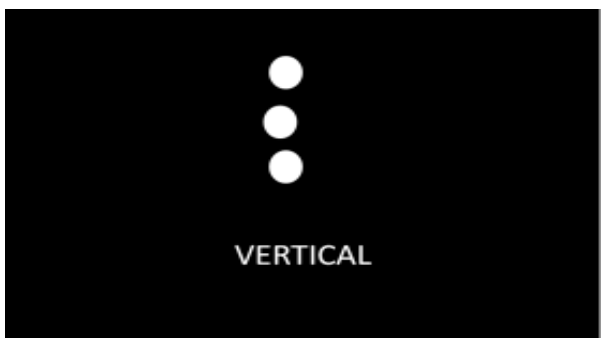

**Figure 7. Vertical Alignment**

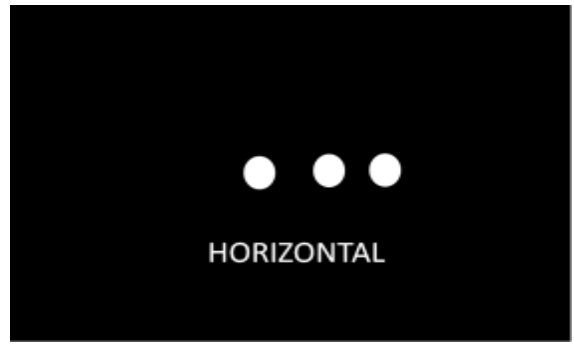

**Figure 8. Horizontal Alignment**

But since we are dealing with objects and not points, we have introduced a threshold value which when crossed will determine if the horizontal or vertical alignment is present. (**Figure 9**)

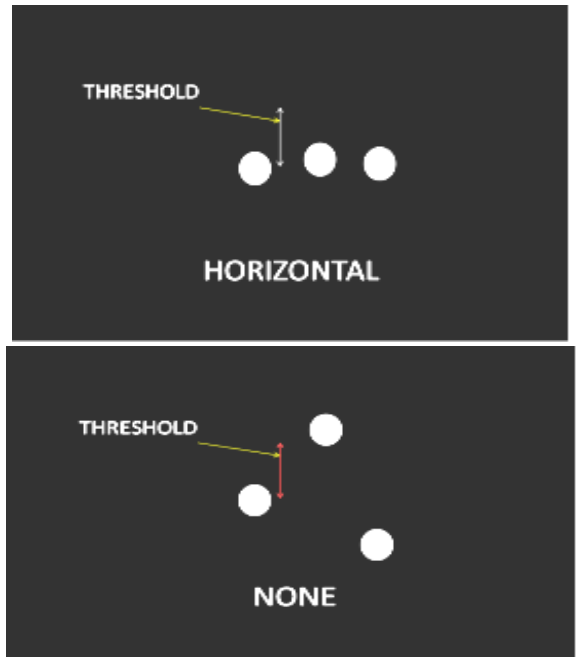

**Figure 9. Breaching of threshold causes ambiguity of alignment**

### 3.4 Direction Detection
Many human gestures have pointing of a finger or two involved. This too is an important gesture which cannot be done away with.
To detect the direction of the finger-point with respect to the fist, we developed a simple algorithm in which the degree of inclination of the finger point from the centre of the palm is determined as shown in **Figure 10** and **Figure 11.**

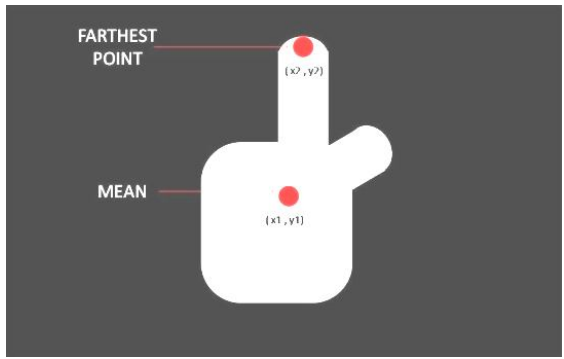

**Figure 10. Mean of the palm and the farthest point from it**

The algorithm is as follows:
Step 1. Find the mean position of all white pixels in the cluster
Step 2. Find the farthest point of that cluster from the mean (**Figure 10**)
Step 3. Calculate the inclination of the line joining the farthest point and the mean, considering the mean as the centre of a 2D coordinate system.
Step 4. End

This algorithm returns the inclination in the range **[0,180]U[0,-180]. (Figure 11)**

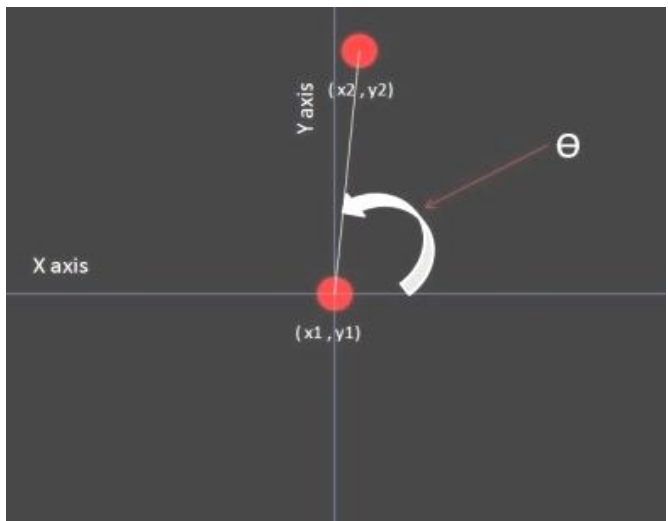

**Figure 11. Angle of inclination θ**

### 3.5 Detect up-or-down Gesture
One can move one's hands up or down and depending on the context; a scroll action can be performed. The algorithm is as follows:

Step 1. Track the position of the clusters in view
Step 2. If the cluster(s) remain stationary for more than the number of frames predetermined by the user, or if they go out of view, then goto step 5.
Step 3. Keep a count of the number of frames and also keep track of the initial and present positions of the cluster(s)
Step 4. Goto step 1
Step 5. If the vertical displacement is less than the threshold, then goto step 8
Step 6. Calculate the speed of motion (as displacement÷number of frames)
Step 7. Return that up-or-down gesture was performed.
Step 8. End

This gesture can now be implemented to perform some action such as mouse wheel up or mouse wheel down actions.
A similar algorithm was used to detect if the user performed left or right action by checking the horizontal displacement instead of the vertical displacement.

### 3.6 Detect left-or-right Gesture
A swipe action is conveniently performed by moving the fingers on a screen from left-to-right or vice-versa.
This too can be converted into a gesture. One can move one's hands left or right and depending on the context, a swipe action can be performed. The algorithm is as follows:

Step 1. Track the position of the clusters in view
Step 2. If the cluster(s) remain stationary for more than the number of frames predetermined by the user, or if they go out of view, then goto step 5.
Step 3. Keep a count of the number of frames and also keep track of the initial and present positions of the cluster(s)
Step 4. Goto step 1
Step 5. If the horizontal displacement is less than the threshold, then goto step 8
Step 6. Calculate the speed of motion (as displacement÷number of frames)
Step 7. Return that left-or-right gesture was performed.
Step 8. End

### 3.7 Detect in-out Gesture
In our world, presently, zooming in/out is a very common gesture. Spreading out our fingers is usually a zoom-in and the reverse is a zoom-out.
We created the following algorithm to detect whether the hand is moving towards the screen (in-action) or away from the screen (out-action).

Step 1. Capture the first frame having the cluster(s). (**Figure 12**)
Step 2. Capture the second consecutive frame having the next changed or unchanged cluster(s). (**Figure 13**)
Step 3. Subtract the first frame from the second. (**Figure 14**)
Step 4. Calculate the sum of number of pixels, of the resulting difference.
Step 5. If the absolute value of the sum is less than the predefined threshold, then goto step 7

Step 6. If sign of the sum is negative, then return zoom-out, else return zoom-in
Step 7. End

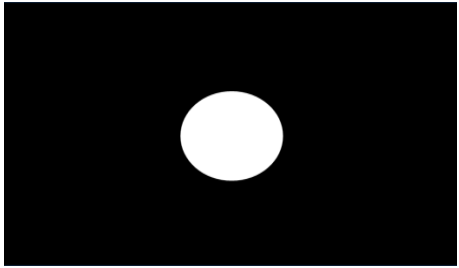

**Figure 12. First Frame**

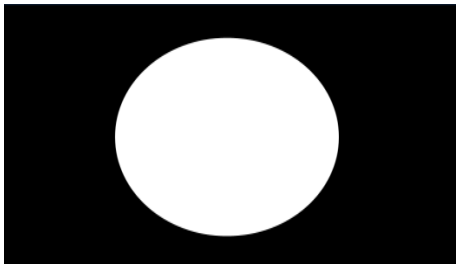

**Figure 13. Second Frame**

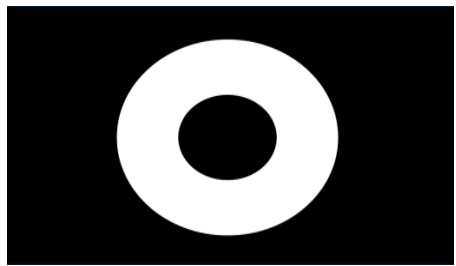

**Figure 14. Frame after subtraction**

### 3.8 Rotation Detection

Circular motion of hands is another common human gesture. It can be in clockwise sense or anti-clockwise sense. To detect a rotation we have followed the following concept:
The general equation of a circle in 2D system is

$$x^2+y^2+Cx+Dy+E=0$$

So, we require three points to find the three unknowns (C,D,E).
For this we capture three consecutive frames.
We find the mean of the top-left white cluster and store it as a point, for all the three frames, respectively.
We now get an equation of the form **AX=B,** where X=[C,D,E]; A has coefficients of C, D and E; B contains the constant terms. Solving for X we get values of C, D, and E.
Using these values we get the centre and radius of the circle as follows:
**Centre = [–C/2, – D/2] and Radius = ((C²+D²) / 4 – E) ⁰·⁵**

From here we can use the value of **Radius** to check it with the threshold given by the programmer.

Having done the above we now need to detect clockwise and anti-clockwise rotation.
For this we first find the slope of the line joining the first two points. Suppose it is **m**. We rotate the system by an angle **(-θ)**, where **θ =tan⁻¹(m).**
Thus, after normalizing for different orientations in the plane we now just have to deal with four cases (**Figure 15**).

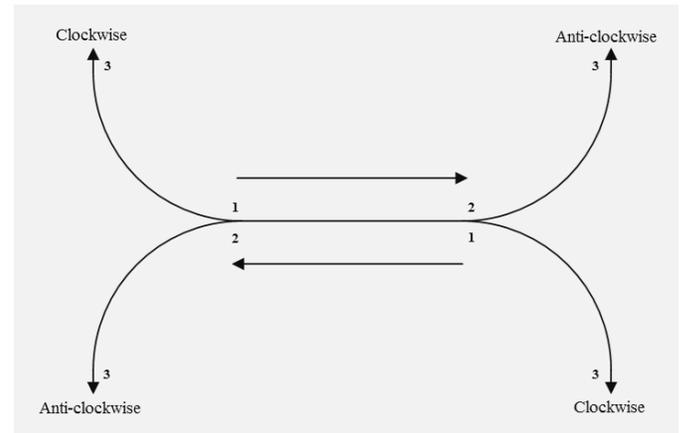

**Figure 15. The four cases for detecting rotation**

when $x_1<x_2$ and $y_3>y_2$ then it is anti-clockwise
when $x_1<x_2$ and $y_3<y_2$ then it is clockwise
when $x_1>x_2$ and $y_3>y_2$ then it is clockwise
when $x_1>x_2$ and $y_3<y_2$ then it is anti-clockwise

Hence in this way we get the type of rotation the user has performed.

### 4. RESULT AND APPLICATION

To test our library functions, we created a **demo application** using MATLAB R2013a.
We assigned a red color to our fingertips using red LEDs (**Figure 16, Figure 17**). The RGB range of the red LEDs were given as input to the functions so that only the fingertips could be recognized by the system. All the functionalities of the computer mouse were implemented using this library. **Figure 18**, **Figure 19** and **Figure 20** show three common mouse functionalities.

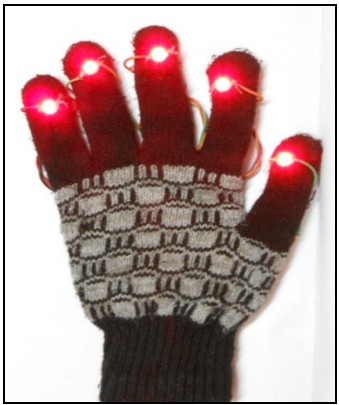

**Figure 16. A glove used to assign red color to the fingertips using LEDs**

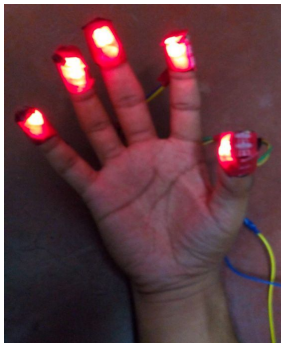

**Figure 17. Finger-caps used to assign red color to the fingertips using LEDs**

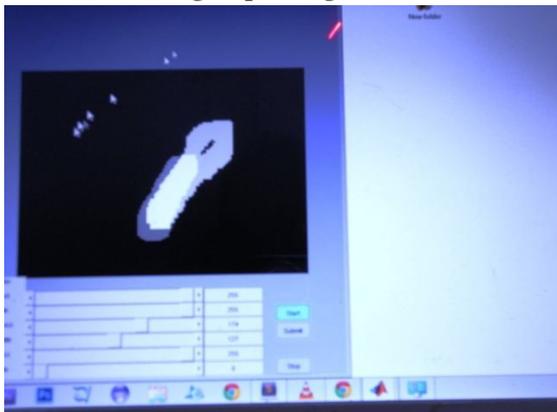

**Figure 18. Implementation of Mouse-movement (using 1 finger)**

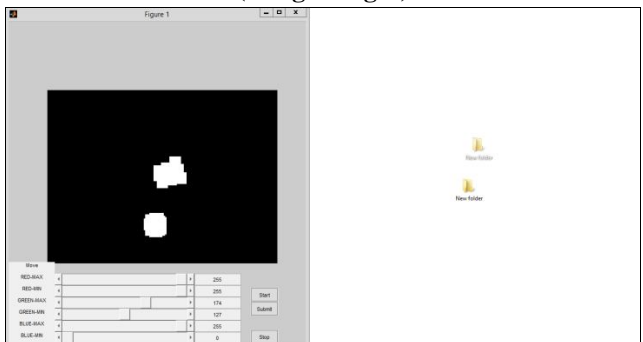

**Figure 19. Implementation of Drag-and-Drop (using 2 fingers)**

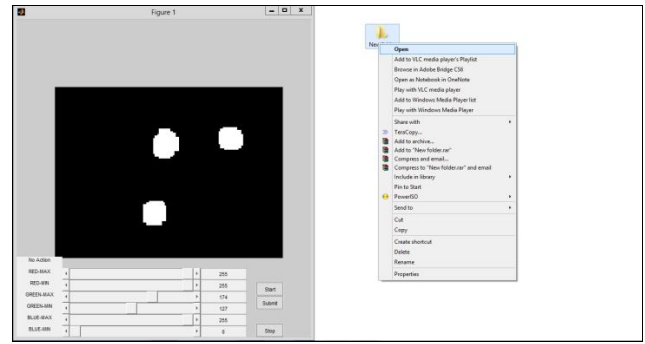

**Figure 20. Implementation of Right-Click (using 3 fingers)**

## 5. CONCLUSION AND FUTURE WORKS

The **Hand Gesture Recognition Library** has thus been able to efficiently recognize most commonly used human gestures. It is not an application and thus can be used by any developer and customized to any extent. It thus enables the developers to utilize the library functions with great flexibility.

This library has been developed using the **MATLAB (R2013a)** environment. In the future it can be modified to run in any other environment. Also, more complicated gestures can be added to the library, such as recognizing the finger type (such as index-finger or thumb).